\documentclass[10pt,final,conference]{IEEEtran}
\usepackage{dcolumn}        % improved alignment of table columns
\usepackage{booktabs}       % improved horizontal lines in tables, toprule
\usepackage{multirow}
\usepackage{makecell}
\usepackage[font=small]{caption} % for // newlines in captions, fixes table cap

\usepackage{cite}
\ifCLASSINFOpdf
  \usepackage[pdftex]{graphicx}
  \graphicspath{{img/}{../pdf/}{../jpeg/}}
  \DeclareGraphicsExtensions{.pdf,.jpeg,.png}
\else
\fi
\usepackage{amsmath}
\usepackage{amsfonts}       % math fonts
\usepackage{amsthm}         % theorems, definitions, etc.
\theoremstyle{plain}
\newtheorem{definition}{Definition}

\usepackage{fixmath} %mathbold

\usepackage{algorithm}% http://ctan.org/pkg/algorithms
\usepackage{algpseudocode}% http://ctan.org/pkg/algorithmicx

\newcommand{\cd}[1]{\texttt{#1}}

\ifCLASSOPTIONcompsoc
 \usepackage[caption=false,font=normalsize,labelfont=sf,textfont=sf]{subfig}
\else
 \usepackage[caption=false,font=footnotesize]{subfig}
\fi

\usepackage{stfloats}
\fnbelowfloat
\usepackage{url}
\begin{document}
%
% paper title
% Titles are generally capitalized except for words such as a, an, and, as,
% at, but, by, for, in, nor, of, on, or, the, to and up, which are usually
% not capitalized unless they are the first or last word of the title.
% Linebreaks \\ can be used within to get better formatting as desired.
% Do not put math or special symbols in the title.
\title{\mbox{Baselines for Reinforcement Learning in Text Games}}

% author names and affiliations
% use a multiple column layout for up to three different
% affiliations
\author{\IEEEauthorblockN{Mikul\'{a}\v{s} Zelinka}
\IEEEauthorblockA{
Charles University, Faculty of Mathematics and Physics\\
Prague, Czech Republic\\
zelinka@ktiml.mff.cuni.cz}
% \and
% \IEEEauthorblockN{Homer Simpson}
% \IEEEauthorblockA{Twentieth Century Fox\\
% Springfield, USA\\
% Email: homer@thesimpsons.com}
% \and
% \IEEEauthorblockN{James Kirk\\ and Montgomery Scott}
% \IEEEauthorblockA{Starfleet Academy\\
% San Francisco, California 96678--2391\\
% Telephone: (800) 555--1212\\
%Fax: (888) 555--1212}}
}

% conference papers do not typically use \thanks and this command
% is locked out in conference mode. If really needed, such as for
% the acknowledgment of grants, issue a \IEEEoverridecommandlockouts
% after \documentclass

% for over three affiliations, or if they all won't fit within the width
% of the page, use this alternative format:
% 
%\author{\IEEEauthorblockN{Michael Shell\IEEEauthorrefmark{1},
%Homer Simpson\IEEEauthorrefmark{2},
%James Kirk\IEEEauthorrefmark{3}, 
%Montgomery Scott\IEEEauthorrefmark{3} and
%Eldon Tyrell\IEEEauthorrefmark{4}}
%\IEEEauthorblockA{\IEEEauthorrefmark{1}School of Electrical and Computer Engineering\\
%Georgia Institute of Technology,
%Atlanta, Georgia 30332--0250\\ Email: see http://www.michaelshell.org/contact.html}
%\IEEEauthorblockA{\IEEEauthorrefmark{2}Twentieth Century Fox, Springfield, USA\\
%Email: homer@thesimpsons.com}
%\IEEEauthorblockA{\IEEEauthorrefmark{3}Starfleet Academy, San Francisco, California 96678-2391\\
%Telephone: (800) 555--1212, Fax: (888) 555--1212}
%\IEEEauthorblockA{\IEEEauthorrefmark{4}Tyrell Inc., 123 Replicant Street, Los Angeles, California 90210--4321}}

% use for special paper notices
%\IEEEspecialpapernotice{(Invited Paper)}

\maketitle

% As a general rule, do not put math, special symbols or citations
% in the abstract
\begin{abstract}
The ability to learn optimal control policies in systems where action space is defined by sentences in natural language would allow many interesting real-world applications such as automatic optimisation of dialogue systems. Text-based games with multiple endings and rewards are a promising platform for this task, since their feedback allows us to employ reinforcement learning techniques to jointly learn text representations and control policies. We argue that the key property of AI agents, especially in the text-games context, is their ability to generalise to previously unseen games. We present a minimalistic text-game playing agent, testing its generalisation and transfer learning performance and showing its ability to play multiple games at once. We also present pyfiction, an open-source library for universal access to different text games that could, together with our agent that implements its interface, serve as a baseline for future research.
\end{abstract}

% no keywords

% For peer review papers, you can put extra information on the cover
% page as needed:
% \ifCLASSOPTIONpeerreview
% \begin{center} \bfseries EDICS Category: 3-BBND \end{center}
% \fi
%
% For peerreview papers, this IEEEtran command inserts a page break and
% creates the second title. It will be ignored for other modes.
\IEEEpeerreviewmaketitle

\section{Introduction}
In text games, which are also known as Interactive Fiction (IF) \cite{Montfort:2005}, the player is given a description of the game state in natural language and then chooses one of the actions which are also given by textual descriptions.
The executed action results in a change of the game state, producing new state description and waiting for the player's input again.
This process repeats until the game is over.

The engine that is responsible for running IF games is called simply a \emph{game simulator} and it provides an interface for player (human or bot) interaction.

\begin{figure*}[!ht]
\centering
\includegraphics[width=\textwidth]{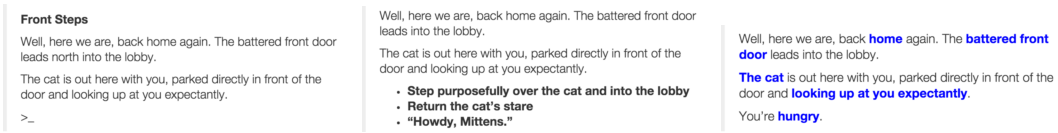}

\caption{Parser, choice and hypertext-based games \cite{DBLP:journals/corr/HeCHGLDO15}.}
\label{fig:if-types}
\end{figure*}

While the output of the game simulator is almost always a text description, the form of the input that the game expects --- the game-player interface --- does vary.
This is one of the most common criteria for classifying IF games (for examples, see Figure \ref{fig:if-types}):

\begin{itemize}

\item \emph{parser-based}, where the player types in any text input freely,

\item \emph{choice-based}, where multiple actions to choose from are typically available, \emph{in~addition} to the state description,

\item \emph{hypertext-based}, where multiple actions are present as clickable links \emph{inside} the state description.
\end{itemize}

The largest source of information on IF games and their variants is the Interactive Fiction Database\footnote{IFDB: \url{http://ifdb.tads.org/}\label{ftn:ifdb}.} which features useful filters, tags and lists for finding specific kinds of games.
For our purposes, though, the differences in game genre or storytelling elements largely do not matter.

Text games come in virtually all genres, often with different minigames or puzzles incorporated into them.
We mostly deal with games without any additional special parts that do not fit into the general and simple input-output loop or we omit them from the selected games.

In this work, we only deal with \emph{choice-based} and \emph{hypertext-based} games but both variants are referred to as \emph{choice-based}, as the hyperlinks are simply considered additional choices.
In fact, all parser-based games with finite number of actions can be converted to choice-based games by simply enumerating all the actions accepted by the interpreter at different time steps.
This trick was employed by \cite{DBLP:journals/corr/NarasimhanKB15} where a subset of all possible action combinations was presented as choices to the agent.

More formally speaking, text games are sequential decision making tasks with both state and action spaces given in natural language.
In fact, a text game can be seen as a dialogue between the game and the player and the task is to find an optimal policy (a mapping from states to actions) that would maximise the player's total reward.
Consequently, the task of learning to play text games is very similar to learning how to correctly answer in a dialogue, making this task relevant for real-world applications.

Usually, text games have multiple endings and countless paths that the player can take.
It is often clear that some of the paths or endings are better than others and different rewards can be assigned to them.
Availability of these feedback signals makes text games an interesting platform for using reinforcement learning (RL), where one can make use of the feedback provided by the game in order to try and infer the optimal strategy of responding in the given game, and potentially, in a dialogue. 

Our aim is to:

\begin{enumerate}
\item Provide a platform which would enable researchers to easily add new text games. Importantly, this library unifies the interface of different types of text games, enabling researchers to conduct large-scale experiments.
\item Design a minimalistic agent capable of playing basic text games well. This agent will also implement the interface of the text-game library and could be used as a starting point for more complex models in the future.
\end{enumerate}

\section{Related Work}

RL algorithms have a long history of being successfully applied to solving games, for example to Backgammon \cite{DBLP:journals/cacm/Tesauro95} or the more recent general AlphaZero algorithm that achieved superhuman performance in Chess and Go \cite{DBLP:journals/corr/abs-1712-01815}.

There have also been successful attempts \cite{DBLP:journals/corr/NarasimhanKB15},  \cite{DBLP:journals/corr/HeCHGLDO15} at playing IF games using RL agents.
However, the selected games were quite limited in terms of their difficulty and even more importantly, the resulting models had mostly not been tested on games that had not been seen during learning.

While being able to learn to play a text game is a success in itself, the resulting model must generalise to previously unseen data in order to be useful.
In other words, we can merely hypothesise that a successful IF game agent can at least partly understand the underlying game state and potentially transfer the knowledge to other, previously unseen, games, or even natural dialogues.
And for the most part, it remains to be seen how the RL agents presented in \cite{DBLP:journals/corr/NarasimhanKB15} and \cite{DBLP:journals/corr/HeCHGLDO15} perform in terms of generalisation in the domain of IF games.

Prior relevant work includes learning to play the game Civilization by leveraging text manuals \cite{DBLP:journals/corr/BranavanSB14} and achieving human-like performance on Atari video games \cite{DBLP:journals/nature/MnihKSRVBGRFOPB15}.

As our agent is partly inspired by the architectures employed in \cite{DBLP:journals/corr/NarasimhanKB15} and \cite{DBLP:journals/corr/HeCHGLDO15}, these models are next described in more detail.

The LSTM-DQN \cite{DBLP:journals/corr/NarasimhanKB15} agent uses an LSTM network \cite{hochreiter1991untersuchungen} for representation generation, followed by a variant of a Deep Q-Network (DQN) \cite{DBLP:journals/nature/MnihKSRVBGRFOPB15} used for scoring the generated state and action vectors. The underlying task of playing parser-based games is effectively reduced to playing choice-based games by presenting a subset of all possible verb-object tuples as available actions to the agent.

In the framework, a single action consists of a verb and an object (e.g. \emph{eat apple}) and the model computes Q-values for objects --- $Q(s, o)$ --- and actions --- $Q(s, a)$ --- separately. The final Q-value is obtained by averaging these two values.

The Deep Reinforcement Relevance Network (DRRN) \cite{DBLP:journals/corr/HeCHGLDO15} was introduced for playing hypertext-based games.
The agent learns to play \emph{Saving John} and \emph{Machine of Death} games and used a simple bag-of-words (BOW) representation of the input texts.

The learning algorithm is also a variant of DQN; DRRN refers to the network that estimates the Q-value, using separate embeddings for states and actions. DRRN then uses a variable number of hidden layers and makes use of softmax action-selection.
The final Q-value is obtained by computing an inner product of the inner representation vectors of states and actions.

\section{Background}

We formally state the problem of learning to play the games as solving a Markov decision process.
Then, we briefly review common machine learning methods that will be employed, most notably neural networks and Q-learning.

\subsection{Formulating the learning task}
\label{sec:task}

Text game is a sequential decision-making task with both input and output spaces given in natural language.

\begin{definition}[Text game]
\label{def:text-game}
Let us define a text game as a tuple $G = \langle H, H_t, S, A, \mathcal{D}, \mathcal{T}, \mathcal{R} \rangle$, where
\begin{itemize}
\item $H$ is a set of game states, $H_t$ is a set of terminating game states, $H_t \subseteq H$,
\item $S$ is a set of possible state descriptions,
\item $A$ is a set of possible action descriptions,
\item $\mathcal{D}$ is a function generating text descriptions, $\mathcal{D}: H \to (S \times 2^A)$,
\item $\mathcal{T}$ is a transition function, $\mathcal{T}: (H \times A) \to H$,
\item $\mathcal{R}$ is a reward function, $\mathcal{R}: (S_t, A_t, S_{t+1}) \to \mathbb{R}$.
\end{itemize}
\end{definition}

Generally speaking, both the transition function $\mathcal{T}$ and the description function $\mathcal{D}$ may be stochastic and the properties of the game and its functions have a great impact on the task difficulty\footnote{For a more thorough discussion, please refer to \cite{zelinka}.}.

In particular, if both $\mathcal{D}$ and $\mathcal{T}$ are deterministic, the \emph{whole game} is deterministic.
Consequently the problem is then reduced to a simple graph search problem that can be solved by graph search techniques and there is no need --- from the perspective of finding optimal rewards --- to employ reinforcement learning methods.
However, from the generalisation perspective, it still might be useful to attempt to learn simple games using RL techniques as the agents could potentially be able to generalise or transfer their knowledge to other, more difficult problems.

We define the task of learning to play a text game as the process of finding an optimal policy for the respective text game Markov decision process (MDP) \cite{DBLP:books/lib/SuttonB98}, where $\mathcal{M}~=~\langle S_\mathcal{M}, A_\mathcal{M}, \mathcal{P}, \mathcal{R}, \gamma_\mathcal{M} \rangle$.

States, actions and the reward function $S$, $A$, $\mathcal{R}$ of the game correspond to the MDP states, actions and the reward function  $S_\mathcal{M}$, $A_\mathcal{M}$, $\mathcal{R}$.
The MDP probability function $\mathcal{P}(s' | s, a)$ is realised by the game transition function $\mathcal{T}$ and is crucially unknown to the agent.

Notice that not all IF games necessarily have the Markov property, i.e. in text games, there can be long-term dependencies.
In fact, it is even difficult to determine whether a text game is or is not Markovian.
However, even problems with non-Markovian characteristics are commonly represented and modelled as MDPs while still giving good results \cite{DBLP:books/lib/SuttonB98}.

Following the common notation, the agent tries to maximise the  \textbf{discounted cumulative reward} $\sum^{\infty}_{t=0} {\gamma^t \mathcal{R} (s_t, a_t, s_{t+1})}$ while following the policy $\pi(s)$ by selecting actions $a \gets \pi(s)$.
The goal is to learn the \textbf{optimal policy $\pi^*$} by maximising the following term:

\begin{equation}
\label{eq:optimal-policy}
\sum^{\infty}_{t=0} {\gamma^t \mathcal{R} (s_t, a_t, s_{t+1})}\text{, where } a_t = \pi^{*}(s_t).
\end{equation}

Using this approach, we have now formally reduced the task of learning to play the text-based games to estimating an optimal policy in a text-game MDP.
The Q-learning algorithm and its neural network estimator for solving the problem are described next.

\subsection{Deep Reinforcement Learning}

For finding the optimal policy in the text-game MDP, we employ Deep Reinforcement Learning (DRL) \cite{DBLP:journals/nature/MnihKSRVBGRFOPB15}. This term is commonly used to describe RL algorithms that use (deep) neural networks as a part of their value function estimation.

Reinforcement learning is an area of machine learning methods and problems based on the notion of learning from numerical rewards obtained by an agent through interacting with an environment \cite{DBLP:books/lib/SuttonB98}.

\begin{figure}[!ht]\centering
\includegraphics[]{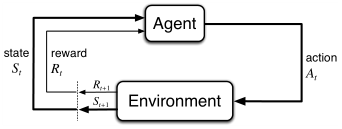}
\caption{Interaction between the agent and the environment \cite{DBLP:books/lib/SuttonB98}.}
\label{fig:rl-loop}
\end{figure}

In any given step, the agent observes a \emph{state} of the environment and receives a~\emph{reward signal}.
Based on the current state and the agent's behaviour function --- the \emph{policy} --- the~agent chooses an~\emph{action} to take.
The action is then sent to the environment which is updated and the loop repeats. See Figure \ref{fig:rl-loop} for illustration of the agent-environment interface.

There are different approaches to learning the optimal policy.
Here, we focus on a method for model-free control that has been shown to perform well on a variety of game-related tasks \cite{DBLP:journals/cacm/Tesauro95}, \cite{DBLP:journals/nature/MnihKSRVBGRFOPB15},  Q-learning.

Q-learning makes use of the concept of a \emph{value function} which evaluates how good a given state under a given policy is --- what reward we can expect to obtain in the long run if we follow the policy from the specific state.
Action value function that determines the value of taking action $a$ in state $s$ using policy $\pi$ is defined as $Q^\pi(s, a) = \mathbb{E}_{\pi} \left[ \sum_{t=0}^{\infty} \gamma^t r_{t+1}\ \middle|\ s, a \right]$.

% \begin{equation}
% \label{eq:action-value-function}
% Q^\pi(s, a) = \mathbb{E}_{\pi} \left[ \sum_{t=0}^{\infty} \gamma^t r_{t+1}\ \middle|\ s, a \right].
% \end{equation}

Now the optimal policy, denoted $\pi^*$, can be also characterised by the optimal action-value function $Q^*(s, a) = \underset{\pi}{\max}\ Q^\pi(s,a)$.

% \begin{equation}
% \label{eq:optimal-action-value}
% Q^*(s, a) = \underset{\pi}{\max}\ Q^\pi(s,a).
% \end{equation}

In other words, if we know the optimal action-value function $Q^*$, we can obtain the optimal policy $\pi^*$ by simply choosing the actions with maximum Q-values: $\pi^{*}(s) = \underset{a}{\max}\ Q^{*}(s,a)$.

% \begin{equation}
% \label{eq:optimal-q-policy}
% \pi^{*}(s) = \underset{a}{\max}\ Q^{*}(s,a).
% \end{equation}

Typically, the agent's policy is also $\epsilon$-greedy with relation to the Q-function, where $0 \leq \epsilon \leq 1$ is a parameter that corresponds to the probability of choosing a random action.

Since Q-learning attempts to find the optimal Q-values which obey the Bellman equation \cite{bellman2013dynamic}, the update rule for improving the estimate of the Q-function is as follows \cite{DBLP:books/lib/SuttonB98}:
\begin{equation} \label{eq:q}
\begin{split}
 Q(s_{t},a_{t}) \gets Q(s_{t},a_{t}) + \phantom{=}\phantom{=}\phantom{=}\phantom{=}\phantom{=}\phantom{=}\phantom{=}\phantom{=}\phantom{=}\phantom{=}\phantom{=}\phantom{=}\phantom{=}\phantom{.}\\
\alpha_{t} \cdot \left( r_{t+1} + \gamma \cdot \max_{a_{t+1}}Q(s_{t+1}, a_{t+1}) - Q(s_{t},a_{t}) \right),
\end{split}
\end{equation}
where $0 \leq \alpha_t \leq 1$ is the learning rate parameter.

The Q-learning algorithm is guaranteed to converge towards the optimal solution \cite{DBLP:books/lib/SuttonB98}. In simpler problems and by default, it is assumed that the Q-values for all state-action pairs are stored in a table.

This approach is, however, not feasible for more complex problems such as the Atari platform games task \cite{DBLP:journals/nature/MnihKSRVBGRFOPB15} or our text-game task, where the state and action spaces are simply too large to store.
In text games, for example, the spaces are infinite.
We address this problem by approximating the optimal Q-function by a function approximator in the form of a neural network. The Q-function is parametrised as
\begin{equation}
\label{eq:nn-q}
Q^*(s_t, a_t) \approx Q(s_t, a_t, \theta_t) = \theta_t(s_t, a_t),
\end{equation}
where the $\theta$ function is realised by a neural network.

The advantage of this non-tabular approach is that even in infinite spaces, the neural network can generalise to previously unobserved inputs and consequently cover the whole search space with reasonable accuracy.
In contrast to linear function approximators, though, non-linear approximators such as neural networks do not guarantee convergence in this context.

\section{Goals}

Our goal is to introduce a minimalistic architecture serving as a proof of concept, with the ability to capture important sentence-level features and ideally capable of reasonable generalisation to previously unseen data.
First, we highlight some of the aspects of the LSTM-DQN and DRRN models that could be improved upon in terms of these requirements.

The main drawback of DRRN is its use of BOW for representation generation.
Consequently, the model is incapable of properly handling state aliasing and differentiating simple yet important nuances in the input, such as the difference between \emph{``There is a treasure chest to your left and a dragon to your right.''} and \emph{``There is a treasure chest to your right and a dragon to your left.''}.

Moreover, %\cite{DBLP:journals/corr/HeCHGLDO15}
He et al. \cite{DBLP:journals/corr/HeCHGLDO15} claim that separate embeddings for state and action spaces lead to faster convergence and to a better solution. However, since both state and action spaces really contain the same data --- at least in most games and especially in hypertext games where actions are a subset of states --- we aim to employ a joint embedding representation of states and actions.

We also believe that a joint representation of states and actions should eventually lead to stronger generalisation capabilities of the model, since such model should be able to transfer knowledge between state and action descriptions as their representation would be shared.

The LSTM-DQN agent, on the other hand, utilises an LSTM network that can theoretically capture more complex sentence properties such as the word order.
However, its architecture only accepts actions consisting of two words.

Additionally, the two action Q-values are finally averaged, which would arguably be problematic if the verbs and objects were not independent.
For example, the value of the verb ``\emph{drink}'' varies highly based on the object; consider the difference between the values of ``\emph{drink}'' when followed by either ``\emph{water}'' or ``\emph{poison}'' objects.

We thus aim to utilise a minimalistic architecture that should:
\begin{itemize}
\item be able to capture dependencies on sentence level such as word order,
\item accept text input of any length for both states and action descriptions,
\item accept and evaluate any number of actions,
\item use a powerful interaction function between states and actions.
\end{itemize}

\section{Methods}
We present the \emph{pyfiction} platform and specify the relevant learning tasks. Then we describe the architecture of our agent capable of learning the games, leveraging the general game interface of the platform. Finally, we describe our agent architecture in detail.

\subsection{Platform}

The \emph{pyfiction}{\footnote{\emph{pyfiction}: \url{https://github.com/MikulasZelinka/pyfiction}.} platform is a library for universal access to different kinds of IF games.
Its interface is identical to the general RL interface (see Figure \ref{fig:rl-loop}).
Currently, it supports the \emph{Glulxe}, \emph{Z-machine} and general HTML simulators.

There are eight games present as of now, however, adding new games is straightforward.
Apart from the game files, it is only necessary to provide a mapping from states to actions and to numerical rewards for the game to be playable by AI agents.

Any agent compatible with the simple RL interface can directly play all games present in \emph{pyfiction}. Pyfiction can also integrate to OpenAI Gym \cite{DBLP:journals/corr/BrockmanCPSSTZ16}.

\subsection{The SSAQN architecture}
\label{sec:slsn}

Our neural network model is inspired by both LSTM-DQN and DRRN. For the sake of clarity, it is referred to as \textbf{SSAQN} (Siamese State-Action Q-Network).

Similarly to \cite{DBLP:journals/corr/NarasimhanKB15} and \cite{DBLP:journals/corr/HeCHGLDO15}, we employ a variant of DQN \cite{DBLP:journals/nature/MnihKSRVBGRFOPB15} with experience replay and prioritised sampling which uses a neural network (SSAQN) for estimating the DQN's Q-function.

SSAQN uses a siamese network architecture \cite{DBLP:conf/nips/BromleyGLSS93}, where two branches --- a state branch and an action branch --- share most of the layers that effectively generate useful representation features.
This is best illustrated by visualising the network's computational graph; see Figure \ref{fig:architecture}.

As we are using a siamese architecture, the weights of the embedding and LSTM layers are shared between state and action data passing through.
States and actions are only differentiated in the dense layers whose outputs are then fed into the similarity interaction function.

The most important differences to LSTM-DQN and DRRN are:

\begin{itemize}
\item the network accepts two text descriptions (state and action) of arbitrary length as an input,
\item the embedding and LSTM layers are the same for states and action, i.e. their weights are shared,
\item the interaction function of inner vectors of states and actions is realised by a normalised dot product, commonly called cosine similarity (see Section \ref{sec:interaction-function}).
\end{itemize}

The output of the SSAQN is the estimated Q-value for the given state-action pair, i.e. the network with parameters $\theta$ realises a function $\theta(s, a) =  Q(s, a, \theta) \approx Q^*(s, a)$ where the input variables $s$ and $a$ contain preprocessed text\footnote{For more details, see the \emph{pyfiction} library.}.

% \begin{equation}
% \theta(s, a) =  Q(s, a, \theta) \approx Q^*(s, a),
% \tag{\ref{eq:nn-q}}
% \end{equation}

To compute the $Q(s, a^i)$ for different $i$ --- for multiple actions --- we simply run the forward pass of the action branch multiple times.

Next, the SSAQN architecture is described layer-by-layer in more detail.

\begin{figure}[!ht]\centering
\includegraphics[width=0.5\textwidth]{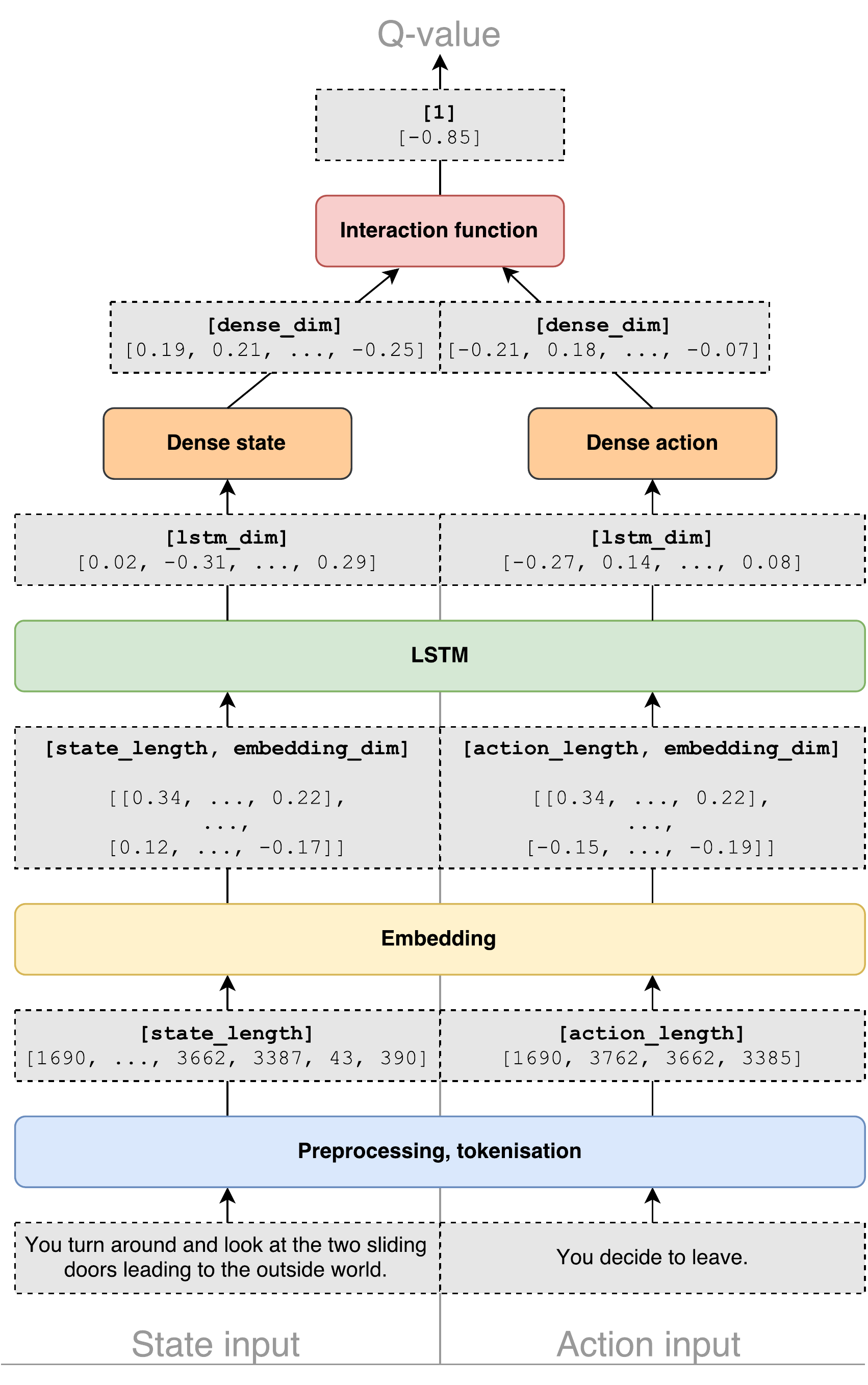}
\caption[Architecture of the SSAQN model and its data flow.]{
Architecture of the SSAQN model and its data flow. Grey boxes represent data values in different layers; the bold text corresponds to the shape of the data~tensors.}
\label{fig:architecture}
\end{figure}

\subsubsection{Word embeddings}
\label{sec:emb-layer}

After preprocessing the text, we convert the words to their vector representation using word embeddings \cite{DBLP:conf/nips/MikolovSCCD13}.
Since our dataset is comparatively small, we use a relatively low dimensionality for the word representations and work with \cd{embedding\_dim} of~$16$.
The weights of the word embeddings are initialised to small random values and trained as a part of the gradient descent algorithm. Using pre-trained vector models is also supported.

As the training is done in batches, we pad the input sequences of words for each batch so that they are all aligned to the same length. Still, note that in general, the length of the output of the embedding layer is variable in length.

\subsubsection{LSTM layer}
\label{sec:lstm-layer}

The inputs of the LSTM layer are the word embedding vectors of variable length.
Similarly to embeddings, the weights of the LSTM units are also initialised to small random values and their weights are shared between states and actions.

The role of the LSTM layer is to successively look at the input words, changing its inner state in the process and thus detecting time-based features in its input.
It is also at this layer that we go from having a data shape of arbitrary length to having a fixed-length output vector.
The output size is equal to the number of LSTM units in the layer and in our experiments, we use \cd{lstm\_dim} of~$32$.

\subsubsection{Dense layer}
\label{sec:dense-layer}

Following the shared LSTM layer, we now have two dense (fully-connected) layers, one for states and one for actions.
Again, we initialise the weights randomly and we also apply the hyperbolic tangent activation function $\mathrm{tanh}(x) = \frac{1 - e^{-2x}} {1 + e^{-2x}}$.

% \begin{equation}
% \label{eq:tanh}
% \mathrm{tanh}(x) = \frac{1 - e^{-2x}} {1 + e^{-2x}}.
% \end{equation}

As the dense layers for states and actions are the only layers to not necessarily share weights between the two network branches, they do play an important role in building differentiated representations for state and action input data.

Note that as the interaction layer uses a dot product, we require both outputs of the dense layers to be of the same dimension and we set \cd{dense\_dim} of both branches to~$8$.
However, theoretically, it would be both possible and interesting to use different layer dimensions for states and actions at this level, as usually, the original state text descriptions carry more information than action descriptions in IF games.
Thus, a possible extension of the network would be to use two or more hidden dense layers in the  state branch of the network and to only reduce the dimension to the action dimension in the last hidden dense state layer.

\subsubsection{Interaction function}
\label{sec:interaction-function}

Lastly, we apply the cosine similarity interaction function to the state and action dense activations, resulting in the final Q-value.
If the input are two vectors $\mathbold{x}$ and $\mathbold{y}$ of~$\cd{dense\_dim} = n$ elements, we define their cosine similarity as a dot product of their L2-normalised (and consequently unit-sized) equivalents:
\begin{equation}
\label{eq:cos}
\mathrm{cs}(\mathbold{x}, \mathbold{y}) = {\frac{\mathbold{x} \cdot \mathbold{y}} {\|\mathbold{x}\|_2 \|\mathbold{y}\|_2}} = \frac{ \sum\limits_{i=1}^{n}{\mathbold{x}_i  \mathbold{y}_i} }{ \sqrt{\sum\limits_{i=1}^{n}{\mathbold{x}_i^2}}  \sqrt{\sum\limits_{i=1}^{n}{\mathbold{y}_i^2}} },
\end{equation}
which corresponds to the cosine of the angle between the input vectors.

Cosine similarity is commonly used for determining document similarity \cite{huang2008similarity}.
Here, we apply it to the two hidden vectors of dense layer values that should meaningfully represent the condensed information that was originally received as a text input by the network and we interpret the resulting value as an indicator of mutual compatibility of the original state-action pair.
Since the range of values of the $\cos$ function is $[-1, 1]$ and since the original rewards that we aim to estimate have arbitrary values, we scale the approximated Q-values accordingly.

\subsubsection{Loss function and gradient descent}

Recall the Q-learning rule (see equation \ref{eq:q}). We define the loss function at time~$t$ as
\begin{equation}
\label{eq:loss}
\mathcal{L}_t = \left( r_{t} + \gamma \cdot \max_{a}Q(s_{t+1}, a) - Q(s_{t},a_{t}) \right)^2,
\end{equation}
which is simply a mean squared error (MSE) of the last estimated Q-value and the target Q-value.

For gradient descent, we make use of the RMSProp optimiser \cite{rmsprop} that has been shown to perform well in numerous practical applications, especially when applied in LSTM networks \cite{DBLP:journals/corr/DauphinVCB15}.

\subsection{Action selection}
\label{sec:action-selection}

Given an SSAQN $\theta$, where $Q(s,a) \gets \theta(s, a)$, the agent selects an action by following the $\epsilon$-greedy policy $\pi_\epsilon(s)$ \cite{DBLP:books/lib/SuttonB98} realised by the following algorithm:

\begin{algorithm}[!ht]
\caption{Action selection}\label{alg:action-selection}
\begin{algorithmic}[1]
\Statex 
\Statex   $\epsilon$: probability of choosing a random action
\Statex   $h(s,a)$: number of times the agent selected $a$ in $s$ in the current run
\Statex 
\Function{Act}{$s$, $actions$, $\epsilon$, $h$}
  \If{$random() < \epsilon$}
    \Return {random action} \EndIf
  \State $q\_values = \theta(s, a_i)$ \textbf{for} $a_i$ in $actions$ \label{row:4}
  \State $q\_values = (q\_values + 1) / 2$ \Comment{from $[-1, 1]$ to~$[0, 1]$}
  \State $q_i = q_i^{h(s, a_i) + 1}$ \textbf{for} $q_i$ in $q\_values$ \Comment{history function} \label{row:history}
  \State $q\_values = (q\_values \cdot 2) - 1$ \Comment{from $[0, 1]$ to~$[-1, 1]$}
%   \State \textbf{return} $a_i$ with $\max q_i$ \label{row:8} \Comment{\footnotemark}
  \State \textbf{return} $a_i$ with $\max q_i$ \label{row:8}

\EndFunction
\end{algorithmic}
\end{algorithm}

% \footnotetext{Lines \ref{row:4} to \ref{row:8} can be formally written as:\\\phantom{XX} \textbf{return} $\underset{a_i \in actions}{\mathrm{argmax}_{a_i}} {((((\theta(s, a_i) + 1)/2) ^ {h(s, a_i)}) \cdot 2) - 1 }$}

The $\epsilon$ is the exploration parameter.
For sampling in training phase (see Algorithm \ref{alg:dqn}), $\epsilon$ is gradually decayed from the starting value of~$1$, i.e. at first, the agent's policy is completely random.
In testing phase, the agent is greedy, i.e. $\epsilon$ is set to~$0$ and the agent always chooses the action with the maximum Q-value for the given state.

The only important difference between the standard $\epsilon$-greedy control algorithm and our action selection policy is that we additionally employ a \emph{history function}, $h(s, a)$.
The scope of the history function is a single run of the agent on a single game, i.e. it is reset every time a game ends.

The function $h(s,a)$ returns a value equal to the number of times the agent selected action $a$ in state $s$ in the current run.
That is, if the agent never selects an action twice in the same state during a run, the history function has no impact on action selection --- and it penalises the already visited state-action pairs, as seen on line \ref{row:history} of Algorithm \ref{alg:action-selection}.

The history function serves as a very simple form of intrinsic motivation \cite{DBLP:conf/nips/SinghBC04}.
It is similar to optimistic initialisation \cite{DBLP:books/lib/SuttonB98} in that it leads the agent to select previously unexplored state-action tuples.
Additionally, note that the history function is not Markovian in the sense that it takes the whole game episode into account.
In practice, the history function greatly helps the agent to avoid infinite loops, since for many games, it is likely to get stuck in an infinite loop when following a randomly chosen deterministic Markovian policy.

\subsection{Training loop}

Putting together all the parts introduced above, we can now formally describe the agent's learning algorithm.

We use a variant of DQN \cite{DBLP:journals/nature/MnihKSRVBGRFOPB15} with experience replay and prioritised sampling of experience tuples with positive rewards \cite{DBLP:journals/corr/SchaulQAS15}. For more details, see Algorithm \ref{alg:dqn}.

Note that the agent inherently supports playing and learning multiple games at once.

\begin{algorithm}
\caption{Training algorithm (a variant of DQN)}\label{alg:dqn}
\begin{algorithmic}[1]
\Statex 
\Statex   $b$: batch size, $p$: prioritised fraction
\Statex   $\epsilon$: exploration parameter, $\epsilon\_decay$: rate at which $\epsilon$ decreases
\Statex 
\Function{Train}{$episodes$, $b$, $p$, $\epsilon = 1$, $\epsilon\_decay$}

  \State Initialise experience replay memory $\mathcal{D}$
  \State Initialise the neural network $\theta$ with random weights
  \State Initialise all game simulators and load the vocabulary

  \For {$e \in {0, \ldots, episodes - 1}$}
    \State Sample each game once using $\pi_\epsilon$, store experiences into~$\mathcal{D}$
    \State $batch$ $\gets$ $b$ tuples $(s_t, a_t, r_t, s_{t+1}, a_{t+1})$ from $\mathcal{D}$, where a fraction of~$p$ have $r_t > 0$
   
    \For {$i, (s_t^i, a_t^i, r_t^i, s_{t+1}^i, a_{t+1}^i)$ in $batch$}
      \State $target^i \gets r_t^i$
      \If {$a_{t+1}^i$} \Comment{$s_{t+1}^i$ is not terminal}
        \State $ target^i \mathrel{{+}{=}} \gamma  \cdot \mathrm{max}_{a_{t+1}^i}\theta(s_{t+1}^i, a_{t+1}^i)$ 
      \EndIf
    \EndFor
    
    \State Define loss as $\mathcal{L}_e(\theta) \gets (target^i - \theta(s_t^i, a_t^i))^2$ 
    \State Perform gradient descent on $\mathcal{L}_e(\theta)$
    \State $\epsilon \gets \epsilon \cdot \epsilon\_decay$ 
    
  \EndFor
\EndFunction  
\end{algorithmic}
\end{algorithm}

\section{Experiments}

We conduct experiments on six games, \emph{Saving John} (SJ), \emph{Machine of Death} (MoD), \emph{Cat Simulator 2016} (CS), \emph{Star Court} (SC), \emph{The Red Hair} (TRH) and \emph{Transit} (TT).
All of these games are available as a part of \emph{pyfiction}.
Since SJ and MoD are also present in \cite{DBLP:journals/corr/HeCHGLDO15}, results on these two games for both agents are directly comparable.

More details about the games including the annotated endings, are available in the \emph{pyfiction} repository.
For basic statistics, see Table \ref{tab:games-summary}.

\subsection{Setup}

In all experiments, we use the following SSAQN layer dimensions: Embedding: $16$, LSTM: $32$, Dense: $8$. The RMSProp optimiser is used with learning rate between $0.001$ and $0.00001$, batch size of $256$ with the prioritised fraction of $25\%$, $\gamma$ of $0.95$, $\epsilon$ of $1$ and $\epsilon$-decay of $0.99$.
We run each experiment five times.

The different testing scenarios are as follows.

\subsubsection{Single-game task} The agent is trained and evaluated on the same game.
\subsubsection{Transfer learning} We pre-train the agent on five games except the one it is then trained and evaluated on. 
\subsubsection{Generalisation} Same as transfer learning, except the agent is not trained on the final game but only evaluated on the game instead.
\subsubsection{Playing multiple games at once} We train and evaluate a single instance of the agent on all six games at once. In each training step, all games are presented successively to the agent.

In all of the scenarios, the agents are mainly compared to the random agent baseline.
For SJ and MoD, we can also compare the results with the DRRN agent as well as the baselines given in \cite{DBLP:journals/corr/HeCHGLDO15}.

\subsection{Results}

Final rewards from all tasks can be seen in Table \ref{tab:games-summary} and the learning progress is depicted in Figure \ref{fig:results} for most of the tasks.

In the single-game task, the SSAQN agent learns to play all deterministic games optimally.
MoD is not learned optimally, but the agent outperforms the DRRN agent (see Figure \ref{fig:comparison_drrn}).
The agent doesn't significantly outperform the random baseline on SC, however, it is unknown if it is possible to do so.

\begin{figure}[ht!]
  \centering
  
  \subfloat[DRRN on SJ \cite{DBLP:journals/corr/HeCHGLDO15}]{\includegraphics[width=0.25\textwidth]{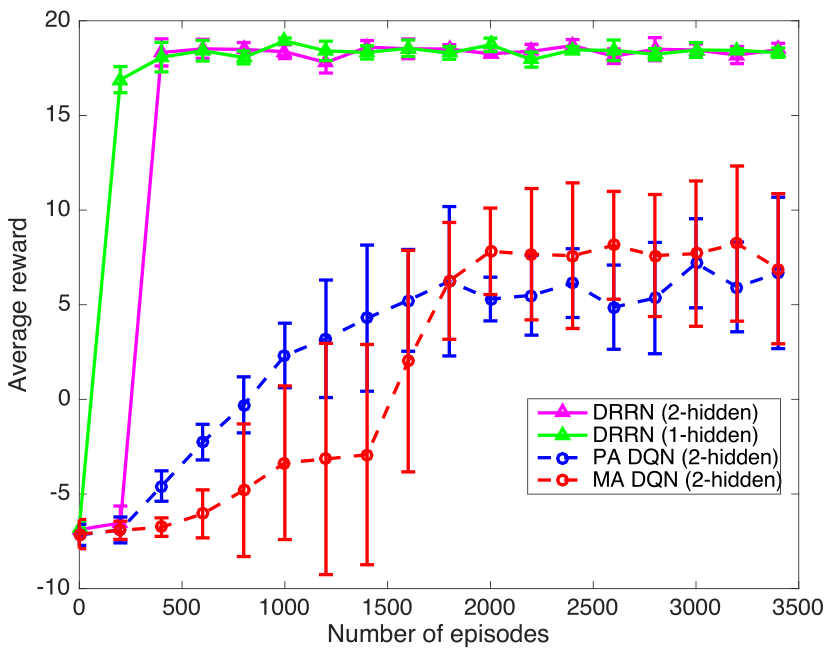}\label{fig:drrn_sj}}
%   \hfill
  \subfloat[DRRN on MoD \cite{DBLP:journals/corr/HeCHGLDO15}]{\includegraphics[width=0.25\textwidth]{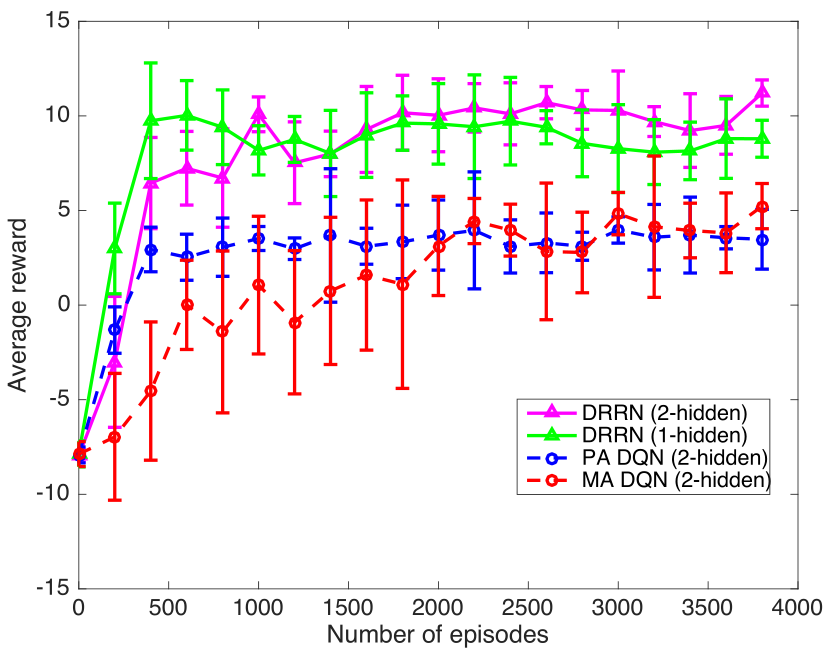}\label{fig:drrn_mod}}

  \subfloat[SSAQN on SJ]{\includegraphics[width=0.25\textwidth]{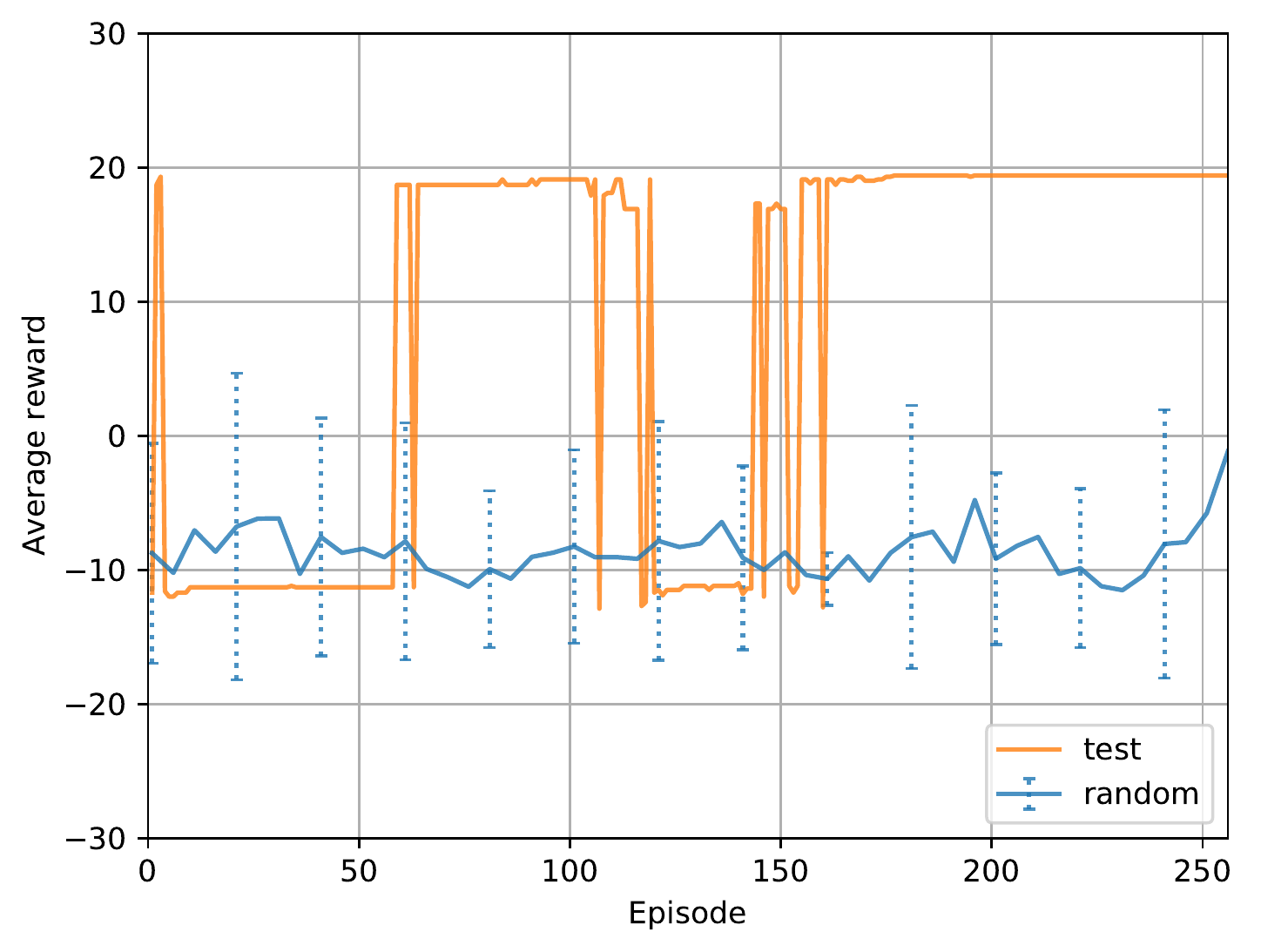}\label{fig:random_sj}}
%   \hfill
  \subfloat[SSAQN on MoD]{\includegraphics[width=0.25\textwidth]{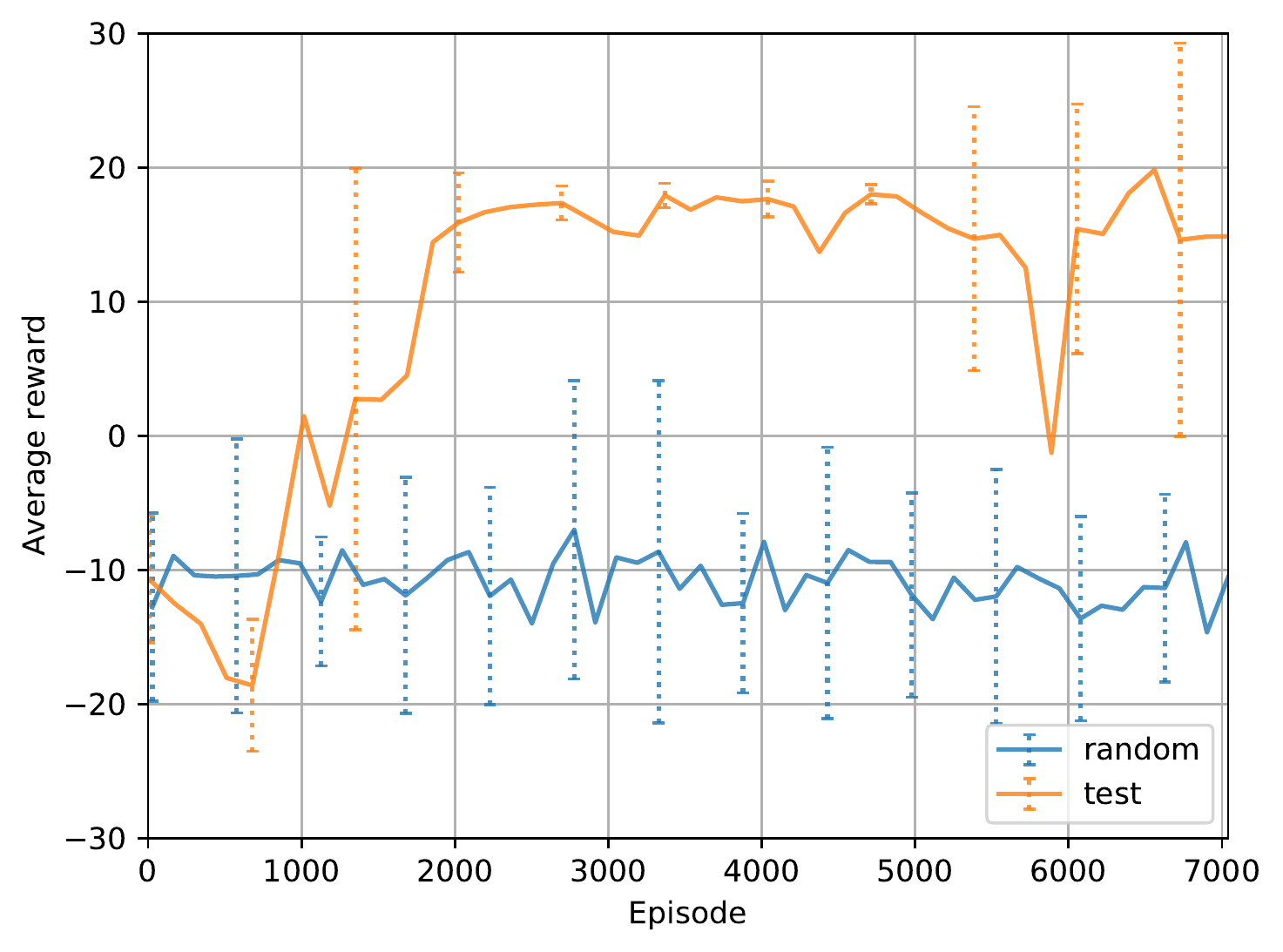}\label{fig:random_mod}}
  
\caption{Comparison of the DRRN and the SSAQN agent Saving John (left) and Machine of Death (right). SSAQN converges considerably faster on both games and achieves higher performance but it has higher variance on MoD. }
  \label{fig:comparison_drrn}
\end{figure}

Transfer learning resulted in a slower convergence rate but similar results on deterministic games and slightly worse results on MoD and SC.
The agent unfortunately but unsurprisingly didn't generalise to unseen games in the generalisation task and a much larger scale of experiments would be needed to safely conclude if it's capable of doing so.
We attribute it mostly to the lack of relevant words and formulations between the games.
Table \ref{tab:games-summary} also depicts how many words from a given game are also present in other games.

Even if this percentage is comparatively high, note that it is just a statistic of single words, meaning it is very unlikely that a word phrases, which are actually important, match between the games.
At any rate, more experiments in different scales are needed to verify this hypothesis.

In the multiple-games tasks, the agent also learned deterministic games optimally, this time however, it was able to also play MoD in an almost optimal way, significantly outperforming DRRN.
Curiously, the result on MoD is better than in the single-game setting, suggesting that information learned from other games might have been useful.

\begin{figure*}[!ht]
  \centering
  
  \subfloat[Saving John]{\includegraphics[width=0.33\textwidth]{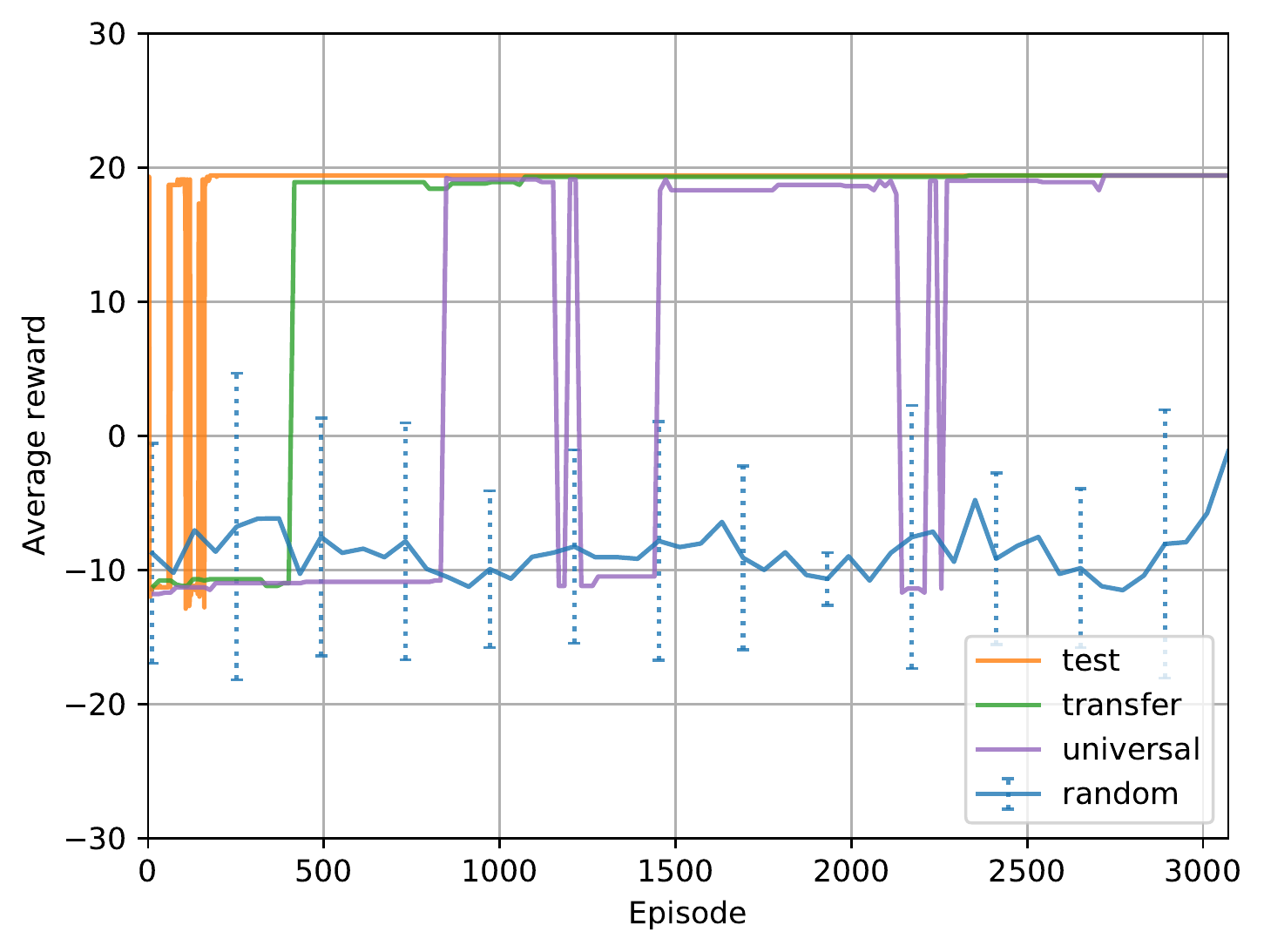}\label{fig:sj}}
  \hfill
  \subfloat[Machine of Death]{\includegraphics[width=0.33\textwidth]{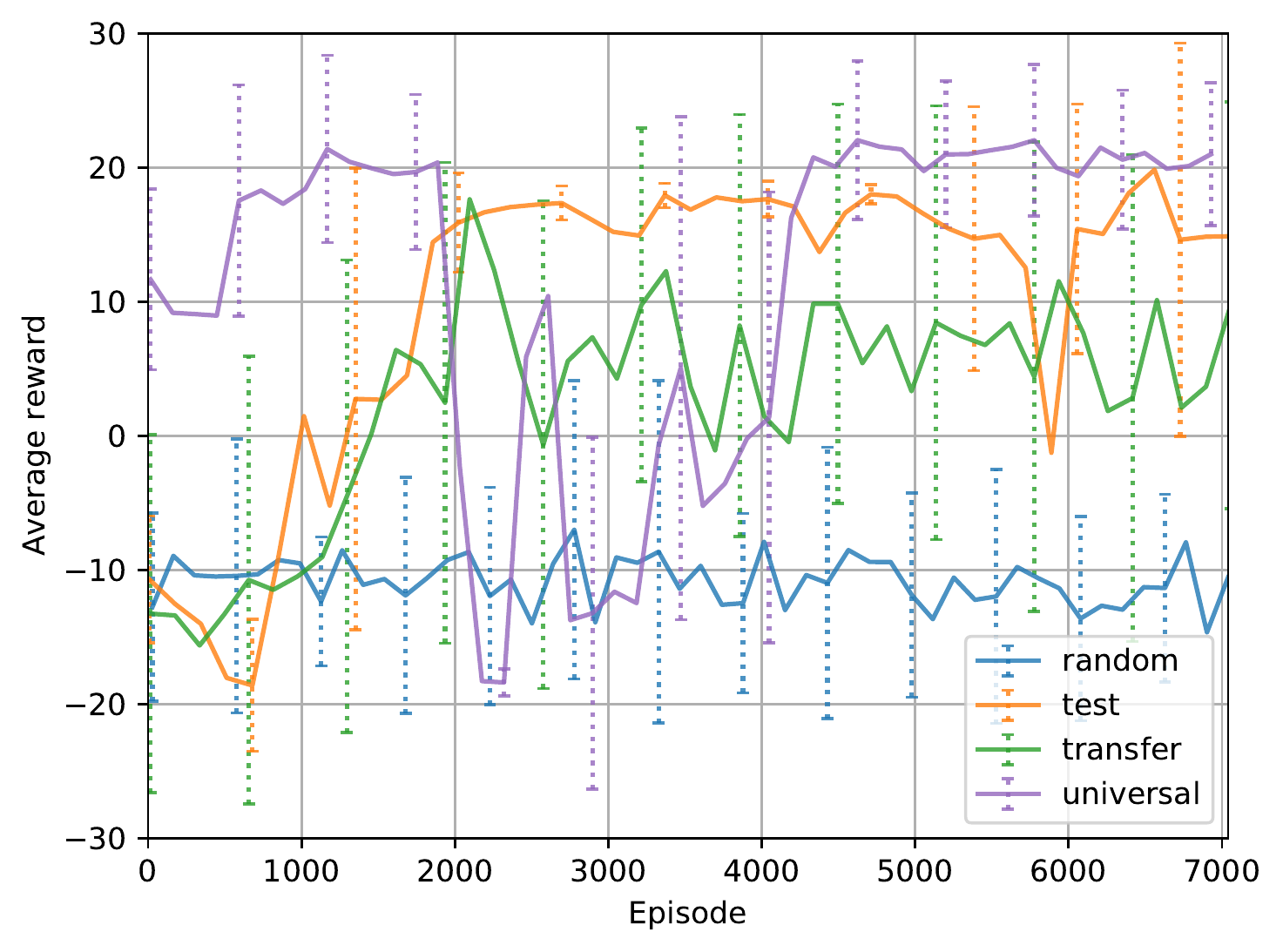}\label{fig:mod}}
  \hfill
    \subfloat[Cat Simulator 2016]{\includegraphics[width=0.33\textwidth]{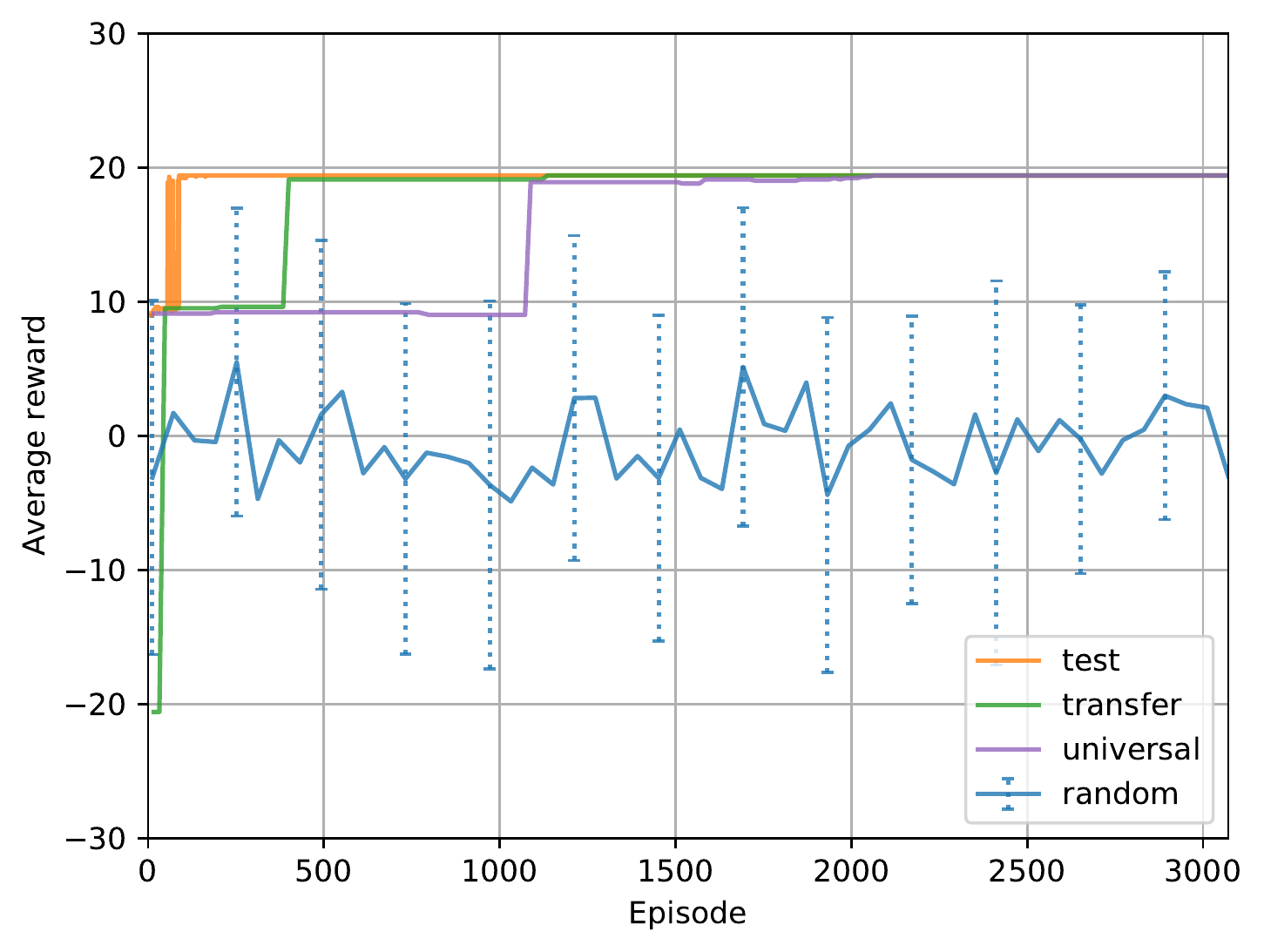}\label{fig:cs}}

  \subfloat[Star Court]{\includegraphics[width=0.33\textwidth]{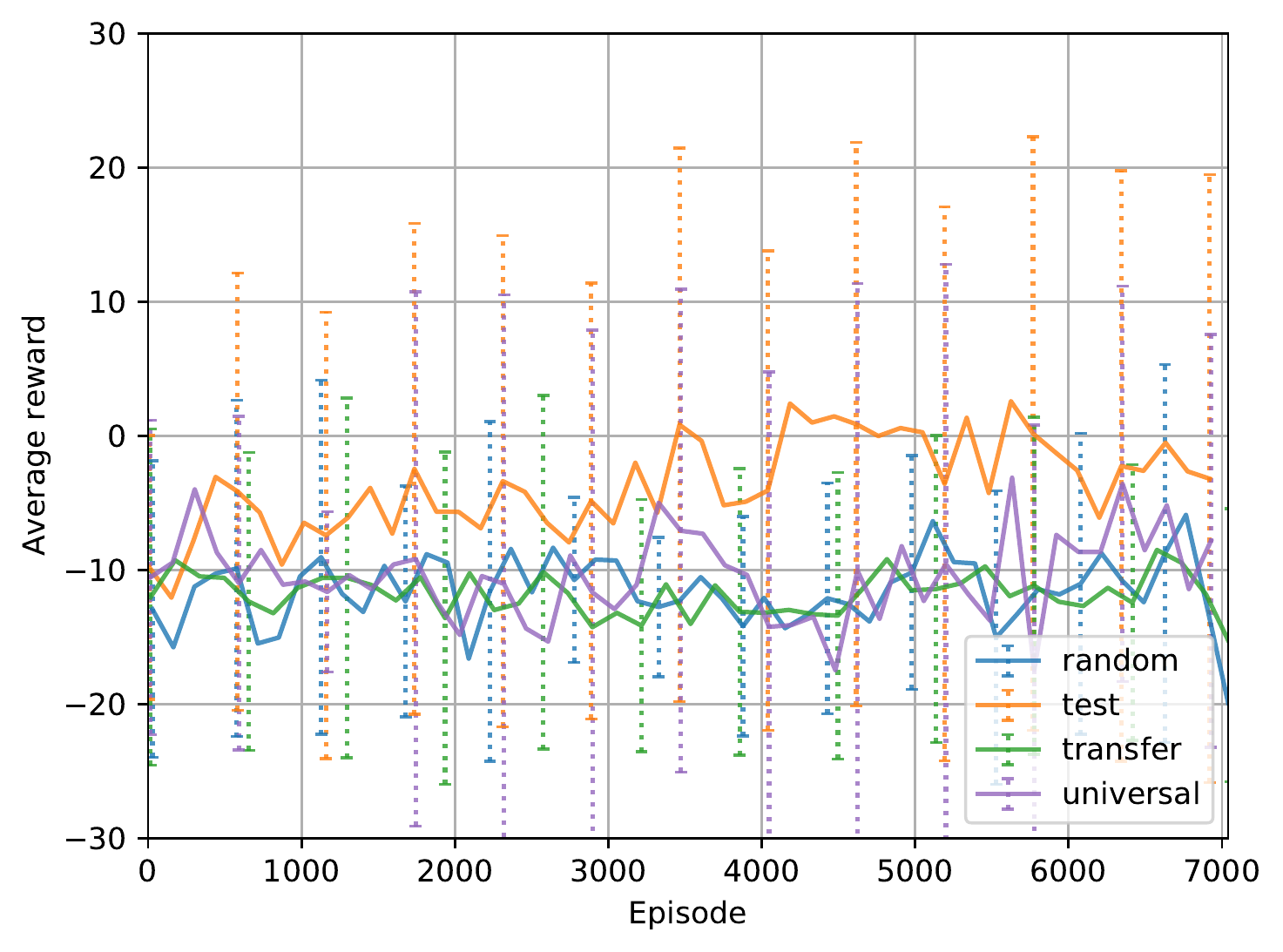}\label{fig:sc}}
  \hfill
  \subfloat[The Red Hair]{\includegraphics[width=0.33\textwidth]{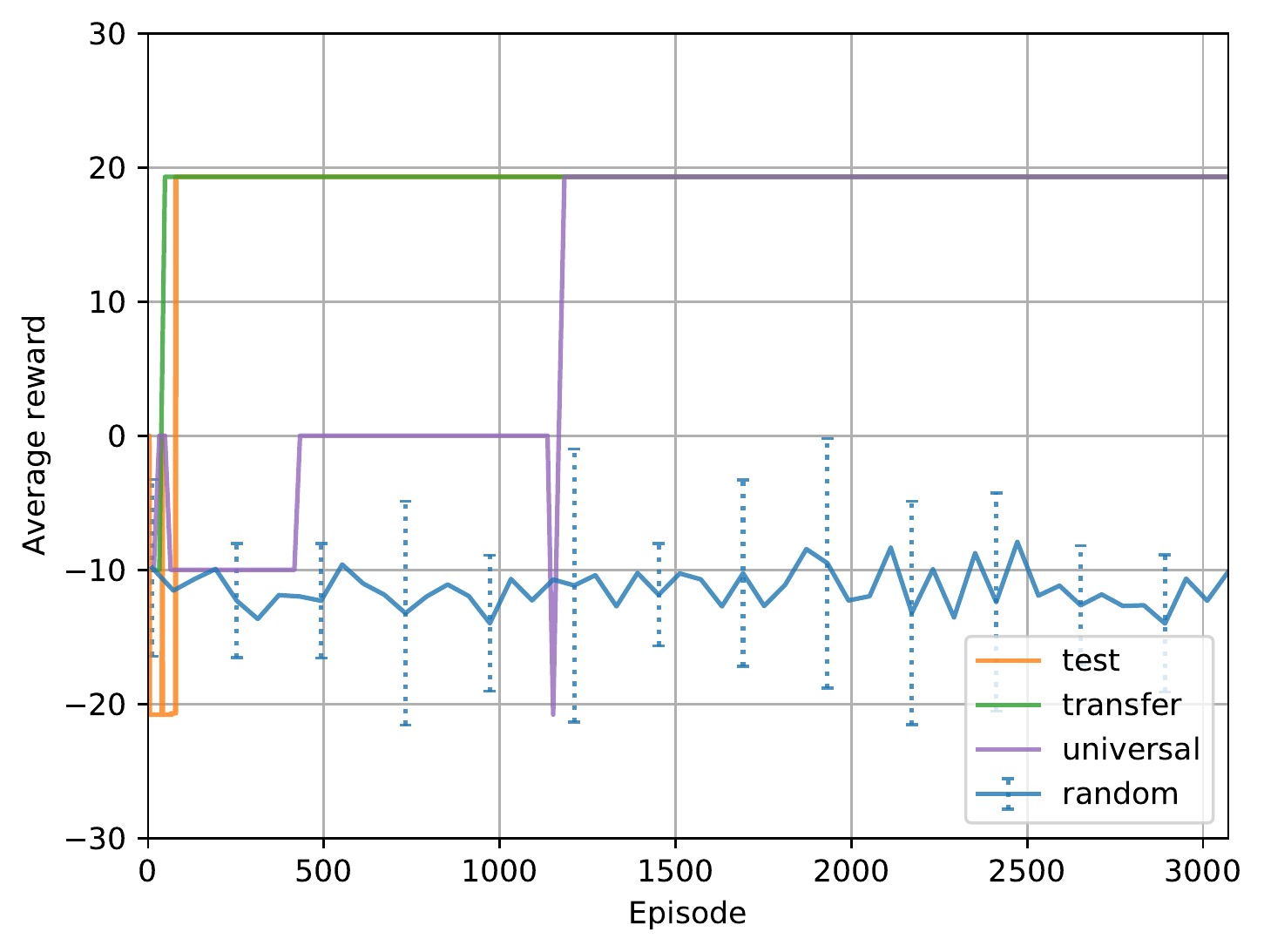}\label{fig:trh}}
  \hfill
    \subfloat[Transit]{\includegraphics[width=0.33\textwidth]{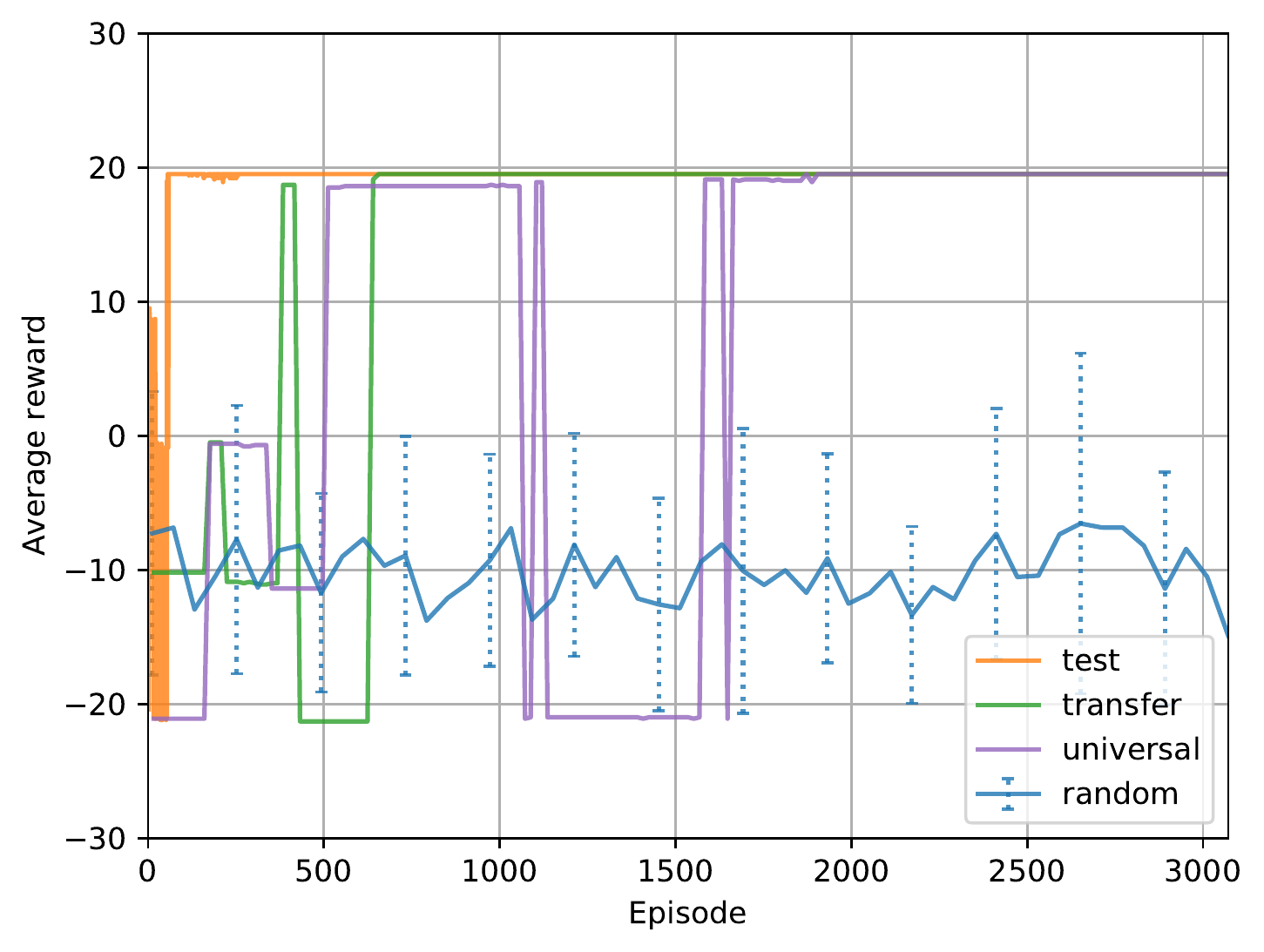}\label{fig:tt}}

  \caption{Results of the SSAQN agent on all six games. Blue: random agent. Orange: train and evaluate on the same game \emph{X}. Green: pre-train on all but \emph{X}, then train and evaluate on \emph{X}. Purple: train and evaluate on all games at once. Standard deviation is shown for the random agent and random games.}
  \label{fig:results}
\end{figure*}

\begin{table*}[!t]
\vspace{2pt}
% \small
\centering
\begin{tabular}{llrrrrrr}
\toprule
& & \textbf{Saving John} & \textbf{Machine of Death} & \textbf{Cat Simulator 2016} & \textbf{Star Court} & \textbf{The Red Hair} & \textbf{Transit} \\

\midrule
\multirow{6}{*}{\rotatebox[origin=c]{90}{\makecell{\textsc{Properties}}}}
                                                               
&\# tokens                             &  1119  &   2055  &   364 &  3929  &  155   &  575  \\
&\# states                             & 70   &  $\geq$ 200   &  37  & $\geq$ 420  &  18   &  76  \\
&\# endings                            &  5  &  $\geq$ 14  &  8  & $\geq$ 17   &    7 &  10  \\
&Avg. words/description                &  73.9  & 71.9    & 74.4   &  66.7  &  28.7   & 87.0   \\
&Deterministic transitions             &  Yes  & No    & Yes   & No   &   Yes  & Yes    \\
&Deterministic descriptions            &   Yes & Yes    &  Yes  &  No  &  Yes   & Yes  \\
&\% of tokens present in other games    &  68.4 & 56.0  & 79.4   &  33.7  &  92.3   & 72.7  \\

\midrule
\multirow{6}{*}{\rotatebox[origin=c]{90}{\makecell{\textsc{Final reward}}}}

& Random agent (average)            &      -8.6       &  -10.8  &      -0.6    &     -11.6  &   -11.4  &     -10.1    \\
& Individual game         &        19.4      & 15.4 &    19.4     &    -2.2    &     19.3   &    19.5     \\
& Generalisation          &         -11.2     & -15.1 &     5.7       &   -13.2 &        -10.0      &    -10.2     \\
& Transfer learning       &           19.4   &   8.7 &        19.4            &     -13.3       &     19.3 &      19.5   \\
& Multiple games           &       19.4      &     21.0       &     19.4       &     -8.2       &      19.3 &     19.5    \\
& Optimal                  &    19.4     &  $\approx$ 21.4  &        19.4       &     ?       &      19.3    &    19.5 \\

\bottomrule
\end{tabular}
\caption[Game properties and agent performance summary.]{Summary of game statistics and performance comparison of the SSAQN agent on different tasks.}
\label{tab:games-summary}
\vspace{3pt}
\end{table*}

\section{Conclusion and Future Work}

We presented a minimalistic text-game playing agent capable of learning to play multiple games at once.
The agent uses a twin-like SSAQN architecture and outperforms the previously suggested DRRN architecture while also being considerably simpler.

We also test the transfer learning and generalisation capabilities of the agent, concluding that it unfortunately doesn't transfer its knowledge or generalise in the limited scope of six games.

To this end, however, we present \emph{pyfiction}, an easily extensible library for universal access to various text games that could, together  with our agent, serve as a baseline for future research.

Future work should mainly focus on expanding the text-game domain and on conducting experiments at a much larger scale.
This should be made much easier thanks to the presented library and we hypothesise that events agents as simple as the SSAQN agent should show some generalisation capabilities given enough data.

% use section* for acknowledegment
\section*{Acknowledgement}

The author would like to thank Rudolf Kadlec for his ideas and guidance during author's work on this topic as a part of Master studies, and to Martin Pil\'{a}t for helpful suggestions and comments.

This research was supported by Charles University under SVV (project number 260 453) and by the Czech Science Foundation (project number 17-17125Y).

% trigger a \newpage just before the given reference
% number - used to balance the columns on the last page
% adjust value as needed - may need to be readjusted if
% the document is modified later
%\IEEEtriggeratref{8}
% The "triggered" command can be changed if desired:
%\IEEEtriggercmd{\enlargethispage{-5in}}

% references section

% can use a bibliography generated by BibTeX as a .bbl file
% BibTeX documentation can be easily obtained at:
% http://mirror.ctan.org/biblio/bibtex/contrib/doc/
% The IEEEtran BibTeX style support page is at:
% http://www.michaelshell.org/tex/ieeetran/bibtex/
\pagebreak
 %
% <OR> manually copy in the resultant .bbl file
% set second argument of \begin to the number of references
% (used to reserve space for the reference number labels box)
% \begin{thebibliography}{1}

% \bibitem{IEEEhowto:kopka}
% H.~Kopka and P.~W. Daly, \emph{A Guide to \LaTeX}, 3rd~ed.\hskip 1em plus
%   0.5em minus 0.4em\relax Harlow, England: Addison-Wesley, 1999.

% \end{thebibliography}

\end{document}